# Towards Sampling from Nondirected Probabilistic Graphical models using a D-Wave Quantum Annealer.

Yaroslav Koshka, M.A. Novotny

*Abstract*—A D-Wave quantum annealer (QA) having a 2048 qubit lattice, with no missing qubits and couplings, allowed embedding of a complete graph of a Restricted Boltzmann Machine (RBM). A handwritten digit OptDigits data set having 8x7 pixels of visible units was used to train the RBM using a classical Contrastive Divergence. Embedding of the classically-trained RBM into the D-Wave lattice was used to demonstrate that the QA offers a high-efficiency alternative to the classical Markov Chain Monte Carlo (MCMC) for reconstructing missing labels of the test images as well as a generative model. At any training iteration, the D-Wave-based classification had classification error more than two times lower than MCMC. The main goal of this study was to investigate the quality of the sample from the RBM model distribution and its comparison to a classical MCMC sample. For the OptDigits dataset, the states in the D-Wave sample belonged to about two times more local valleys compared to the MCMC sample. All the lowest-energy (the highest joint probability) local minima in the MCMC sample were also found by the D-Wave. The D-Wave missed many of the higher-energy local valleys, while finding many "new" local valleys consistently missed by the MCMC. It was established that the "new" local valleys that the D-Wave finds are important for the model distribution in terms of the energy of the corresponding local minima, the width of the local valleys, and the height of the escape barrier.

## I. Introduction

Generative models based on nondirected probabilistic graphical models (Markov Random Fields) [1]-[3] remain very interesting, in part due to their ability to learn and model very complex probability distributions, as well as their suitability for deep unsupervised learning of the distributions represented by training data deprived of classification labels. However, the now classical training algorithms for nondirected models, which include Markov Chain Monte Carlo (MCMC) sampling and other approximations, lately surrendered a big part of their popularity to other models utilizing backpropagation, which proved to be more scalable and therefore were strongly empowered by the improvements in the computational power in the past decades. Besides the inferior scalability of the learning approaches used for nondirected models, there still remain concerns about not only how fast but also how close the model distribution $P_m(v)$ of Markov Random Fields can in practice approach the target distribution $P(v)$ represented by the training samples. Limitations of classical sampling approaches make training generative models very difficult [4].

Back-propagation has been used as a powerful addition to Deep Boltzmann Machines (DBM) [3]. It can be applied as the last step of training, following the DBM pretraining by one of classical techniques aiming at minimizing the Kullback-Leibler-divergence (KL-divergence). However, besides supplementing the nondirected learning algorithm with backpropagation, there are still expectations that significant improvements to the gradient-descent-based log-likelihood maximization could come from more efficient and more precise sampling techniques that would deliver a better representation of the state of the model distribution under optimization.

Quantum Annealers (QAs) are designed to solve a task that at a first glance has nothing to do with generating a sample from any kind of a model distribution. The main purpose of a QA is to find the global minimum or the ground state (GS) of the energy function of an Ising Spin Glass. While the Ising Spin Glass does represent the family of models to which Boltzmann Machines (BM), including DBM, belong, the task of finding the GS is fundamentally different from sampling from a BM model distribution at a value of temperature parameter $T=1$. However, due to the probabilistic nature of finding the GS by a QA, the solution procedure involves almost instantaneous repetition of a large number of solution attempts (usually from 1,000 to 10,000 repetitions), many of which end up to be not the GS but one of many excited states of the Ising Spin Glass energy function. The almost instantaneous generation of this "sample" stimulates an interest in investigating the possibility of using this set of D-Wave solutions to generate or supplement a sample

*Mississippi State University, Mississippi State, MS 39762, U.S.A.* (e-mail: ykoshka@ece.msstate.edu).
M.A. Novotny is with the *Department of Physics and Astronomy, HPC2 Center for Computational Sciences, Mississippi State University, Mississippi State, MS 39762, U.S.A.* (e-mail: man40@msstate.edu).

from a BM model distribution that would not only follow the Boltzmann distribution at *T=1* but would be more representative of the model distribution compared to samples produced with help of classical sampling techniques (e.g., Gibbs sampling).

The potential and difficulties of applying QA hardware to BM problems have been investigated theoretically [5]. The actual quantum hardware was employed in training deep neural networks, and different ways of embedding neural networks into the QA lattice and sampling from the D-Wave have been reported [6]-[12]. Besides using QAs for sampling, new models and training algorithms such as a Quantum BM have been proposed [13] and applied on the D-Wave [14].

In our previous work [15][16], we have made a first attempt to compare classical samples generated using the Gibbs MCMC sampling and a "sample" produced with a help of the D-Wave QA. Statistical comparison of samples obtained by the Gibbs technique versus 1000 sample states generated from the D-Wave revealed significant differences in the observed outcomes. The D-Wave samples were insufficiently representative of the model distribution, specifically by missing many local valleys of the configuration space found by the Gibbs sampling. On the other hand, the D-Wave demonstrated an ability to find local valleys that were consistently missed by the Gibbs sampling, which could potentially be very interesting for sampling applications. However, a few important questions remained unanswered. Specifically, how important are those "new" local valleys found by the D-Wave but missed by the classical MCMC with respect to the RBM energy (i.e., the probability of the corresponding states), the width of the local valleys (and, possibly, the density of the high-probability states) and the height of the escape barrier.

Furthermore, the results in Ref.[15] were obtained for a relatively simple (and a rather ideal in many ways) training and testing dataset - a toy problem of 8×8 bars-and-stripes (BAS) [17]. It remained unclear if the reported trends are applicable to other, more practical and more realistic, training cases, and if those training cases might reveal any additional differences between the results of sampling using the D-Wave versus the classical MCMC.

Finally, there was a slight concern that the less-than-perfect embedding of the RBM into the D-Wave lattice (caused by the D-Wave hardware limitations [11]) could have been partially responsible for some of the observed trends.

In this work, a more systematic investigation of those critical issues is undertaken. A much more realistic popular in ML OptDigits data set [18], with 8 labels (classes), scaled down to 8x7 pixels – was used. Furthermore, we were lucky to have temporary access to D-Wave hardware that was free from some of the limitations responsible for the non-perfect embedding reported by us in Ref. [11]. We continued relying on a flawed but still very useful criterion for comparing the samples, based on monitoring how many and what kind of local valleys of the RBM energy function the generated samples belong to. Convincing evidence was obtained in support of the previous observation that the D-Wave "sample", while missing many important local valleys represented in the MCMC sample, also finds many very important local valleys that are systematically missed by the MCMC. Opportunities for using the discovered property of the QA in order to achieve the ultimate goal - bring the model distribution $P_m(\mathbf{v})$ of Markov Random Fields much closer than what was possible before to the target distribution $P(\mathbf{v})$, while doing that with much higher computational speed compared to the classical sampling approaches - are discussed.

## II. PROBABILISTIC GRAPHICAL MODELS AND QA

### A. Requirements to sampling used for log-likelihood maximization

KL-divergence is the measure of the difference between the target distribution $P(\mathbf{v})$ represented by the training patterns and the model distribution $P_m(\mathbf{v})$.

$$\boldsymbol{KL}\big(P(\mathbf{v}) \parallel P_m(\mathbf{v})\big) \equiv \langle ln(P(\mathbf{v})/P_m(\mathbf{v}))\rangle_P =$$

$$= \sum_{\mathbf{v}} \big(P(\mathbf{v})ln(P(\mathbf{v})) - P(\mathbf{v})ln(P_m(\mathbf{v}))\big) \quad (1)$$

Where $\langle ... \rangle_P$ represents an expectation value with respect to the target distribution. The first term does not depend on the model. When $D$ represents the data composed of $K$ patterns, and $M$ is the graphical model under the optimization, the second term is the expected log-likelihood $ln\,\mathcal{L}(D:M)$. Let us see how the second term is optimized.

If $D = \{\mathbf{v}_T(1), \mathbf{v}_T(2), ..., \mathbf{v}_T(K)\}$:

$$ln\,\boldsymbol{\mathcal{L}}(D:M) \approx -\frac{1}{K}\sum_{k}^{K} ln\big(P_m(\mathbf{v}_T(k))\big) \quad (2)$$

The approximation contains probabilities specifically for each of the training patterns (the probabilities that we want to maximize during the log-likelihood optimization). It may be useful to look at these probabilities to see how the probabilities for states other than the training patterns enter this expression. Specifically for the Energy-based models:

$$P_m(\mathbf{v}) = \frac{e^{-E_{free,m}(\mathbf{v})}}{Z_m} \qquad E_{free,m}(\mathbf{v}) = -ln\sum_{\mathbf{h}} e^{-E(\mathbf{v},\mathbf{h})}$$

$$-\sum_{\mathbf{v}} P(\mathbf{v}) E_{free,m}(\mathbf{v}) \approx -\frac{1}{K}\sum_{k}^{K} E_{free,m}\big(\mathbf{v}_T(k)\big) =$$

$$= \frac{1}{K}\sum_{k}^{K} ln \sum_{\mathbf{h}} e^{-E(\mathbf{v}_T(k),\mathbf{h})}$$

$$ln\,\boldsymbol{\mathcal{L}}(D:M) = -\langle E_{free,m}(\mathbf{v})\rangle_P - ln(Z_m) \approx$$

$$\approx \frac{1}{K}\sum_{k}^{K} ln \sum_{\mathbf{h}}^{all\,\mathbf{h}} e^{-E(\mathbf{v}_T(k),\mathbf{h})} - ln \sum_{\mathbf{v},\mathbf{h}}^{all} e^{-E(\mathbf{v},\mathbf{h})} \quad (3)$$

Therefore, maximizing $ln\,\mathcal{L}(D:M)$ means not only maximizing $ln\,P_m(\mathbf{v}(k))$ log marginal probability of the training data vectors $\mathbf{v}(k)$ under the model (i.e., minimizing the energy $e^{-E(\mathbf{v}(k),\mathbf{h})}$ of the data under the model), it also means reducing the marginal probability (increasing the energy) of all the other states in the state-space.

While it is not feasible to scan through all the **v, h** combinations in the state space, sampling techniques can give an adequate estimate when optimizing the term containing the partition function, as long as the sampling reflects the most important local valleys, those that include states that contribute the most to the partition function (i.e., valleys having high density of states and low state energies $E(\mathbf{v}, \mathbf{h})$).

The optimization of the log-likelihood is conducted using the gradient ascent of the log-likelihood, which utilizes derivatives of the $\ln \mathcal{L}(D : M)$ with respect to coefficients $\theta$ that are used to parametrize the model. The comment above about the required quality of the sample from the model distribution equally applies to the gradient of the log-likelihood during the gradient ascent.

$$\frac{\partial \ln \mathcal{L}(D:M)}{\partial \theta} = -\frac{1}{K}\sum_{k}^{K} \ln \sum_{\mathbf{h}}^{all\ \mathbf{h}} p(\mathbf{h}|\mathbf{v_T}(k))\frac{\partial E(\mathbf{v_T}(k), h)}{\partial \theta}$$

$$+ \sum_{\mathbf{v},\mathbf{h}}^{all\ \mathbf{v},\mathbf{h}} p(\mathbf{v})p(\mathbf{h}|\mathbf{v})\frac{\partial E(\mathbf{v}, \mathbf{h})}{\partial \theta} \approx$$

$$\approx -\frac{1}{K}\sum_{k}^{K} \ln \sum_{\mathbf{h}} p(\mathbf{h}|\mathbf{v_T}(k))\frac{\partial E(\mathbf{v_T}(k), h)}{\partial \theta} +$$

$$+ \frac{1}{K'}\sum_{k'}^{K'} \sum_{\mathbf{h}} p(\mathbf{h}|\mathbf{v_{Sampled}}(k'))\frac{\partial E(\mathbf{v_{Sampled}}(k'), \mathbf{h})}{\partial \theta}$$

Specifically for the energy function $E(\mathbf{v}, \mathbf{h})$ of the RBM, if we go back to the general expression for the log-gradient, it further simplifies, but summation over all possible values of the visible vector $\mathbf{v}$ remains in the second term. For the gradients of the log-likelihood for one training pattern (the expression under the sum over all the training patterns above):

$$\frac{\partial \ln L(\theta|\mathbf{v}(k))}{\partial \omega_{ij}} = [2p(H_i = 1|\mathbf{v}(k)) - 1]v_j(k)$$

$$- \sum_{\mathbf{v}} p(\mathbf{v})[2p(H_i = 1|\mathbf{v}) - 1]v_j(k)$$

$$\frac{\partial \ln L(\theta|\mathbf{v}(k))}{\partial b_j} = v_j(k) - \sum_{\mathbf{v}} p(\mathbf{v}) v_j(k)$$

$$\frac{\partial \ln L(\theta|\mathbf{v}(k))}{\partial c_i} = [2p(H_i = 1|\mathbf{v}(k)) - 1]$$

$$- \sum_{\mathbf{v}} p(\mathbf{v})[2p(H_i = 1|\mathbf{v}(k)) - 1]$$

Ideally, calculation of these gradients would require summing over all possible values of the visible vector $\mathbf{v}$, which is computationally not feasible. This is what the approximated expressions for the gradient of the log-likelihood looks like for the RBM case:

$$\frac{\partial \ln L(\theta|\mathbf{v}(k))}{\partial \omega_{ij}} = [2p(H_i = 1|\mathbf{v}(k)) - 1]v_j(k) -$$

$$- \frac{1}{K'}\sum_{k'}^{K'}[2p(H_i = 1|\mathbf{v_{Sampled}}(k')) - 1]v_{j\ samled}(k')$$

$$\frac{\partial \ln L(\theta|\mathbf{v}(k))}{\partial b_j} = v_j(k) - \frac{1}{K'}\sum_{k'}^{K'} v_{j\ sampled}(k)$$

$$\frac{\partial \ln L(\theta|\mathbf{v}(k))}{\partial c_i} = [2p(H_i = 1|\mathbf{v}(k)) - 1] -$$

$$- \frac{1}{K'}\sum_{k'}^{K'}[2p(H_i = 1|\mathbf{v_{Sampled}}(k')) - 1]$$

When selecting $K'$ equal to the number of the training patterns $K$, the $1/K'$ factor and the summation disappear in the expression used in the iterative algorithm. Once again, in the three expressions above, an adequate estimate of the second terms is possible if the samples $\mathbf{v_{Sampled}}(k')$ represent at least the most important local valleys. For example, consider a sample missing (not finding) some values of $\mathbf{v}$ belonging to one or a few local valleys in the current model distribution that contribute significantly to $Z_m$ (i.e., valleys having high density of states and low state energies $E(\mathbf{v}, \mathbf{h})$). Consistently missing those valleys by the sample will dictate a wrong trajectory of the log-likelihood gradient descent, resulting in the model distribution after training not maximizing the log-likelihood given the training data and assigning high probabilities to wrong states.

When the classical Gibbs technique is used for sampling, each variable is updated sequentially based on its conditional distribution given the state of all the other variables. For RBM, since all hidden units are connected only to visible units and vice versa, the sequential update looks as the following:

$h_i^{(t)} \sim p(h_i \mid \mathbf{v}^{(t)}),\ \ v_i^{(t+1)} \sim p(v_i \mid \mathbf{h}^{(t)})$.

Specifically, for the RBM energy function:

$$p(H_i^{(t)} = 1|\mathbf{v}^{(t)}) = \sigma\left(2\left(\sum_{j=1}^{m}\omega_{ij}v_j^{(t)} + c_i\right)\right),$$

$$p(V_j^{(t+1)} = 1|\mathbf{h}^{(t)}) = \sigma\left(2\left(\sum_{j=1}^{m}\omega_{ij}h_i^{(t)} + b_j\right)\right),$$

where $\sigma(x) = 1/(1 + e^{-x/T})$.

Limitations of classical sampling approaches make training nondirected graphical models very difficult [4]. High-probability states in the model distribution should be reflected by the sampling thanks to the higher probability of those Monte Carlo jumps that lead towards the states of higher conditional probability at the time $(t+1)$ from the current state at $(t)$. It means that not only do states belonging to wider and deeper local valleys have a higher chance of being sampled (which would be reflective of the true probability distribution), but also important is the sign of the gradient (the slope) of $E(\mathbf{v}, \mathbf{h})$ with respect to the coordinates in the space of the random variables $\mathbf{v}$ and $\mathbf{h}$ in the higher-energy areas surrounding particular local valleys. When the corresponding states have low probability, the $E(\mathbf{v}, \mathbf{h})$ gradient in these areas may have little consequence for the true KL-divergence between the target distribution $P(\mathbf{v})$ represented by the training patterns and the model distribution $P_m(\upsilon)$. But this slope may be critical for how well the particular sampling techniques reflect this difference. In other words, in the absence of a sufficient negative slope of $E(\upsilon, \mathbf{h})$ in a wide proximity of some local valleys, the particular local valleys may be missed by the Gibbs sample, despite some of them possibly having a high density of high-probability states (i.e., deep and wide local valleys that we do not want to be missed by the sample).

### B. Adiabatic Quantum Annealing

Adiabatic QA can be considered as an alternative to the classical SA method for finding the global energy minimum of a spin glass [19]. In a QA, the search for the GS utilizes not only thermal fluctuations over barriers (as in the classical SA algorithm), but also quantum mechanical tunneling through the barriers surrounding local minima. The D-Wave was the first commercial quantum annealer (QA) [20]. The D-Wave architecture is an implementation of an Ising spin-glass model [21]. The qubits correspond to an optimization variables $s_j$. There are couplings between some of the qubits. The weights of the couplings $J_{ij}$ and the bias fields $h_j$ are defined to represent a particular spin glass problem Eq. (4) for which the ground state is to be found. The ground state of the Ising spin glass is the states of the Ising spins that minimizes the energy function Eq. (4) [22]. The physics of the ground state determination by the D-Wave QA has been extensively investigated [23]-[25].

While the main job of a QA is to find the GS, the statistical nature of the search for the GS ensures that a big sample from the underlying model distribution, containing not only the GS but also possibly a very large number of excited states, may be generated almost instantaneously. While it is known that the D-Wave sample does not follow the Boltzmann distribution at the desirable value of the temperature parameter $T=1$, there are efficient ways of converting it to the desirable form of the distribution [9]. For that reason, investigation of fundamental properties of the samples produced by the D-Wave is one of the main goals of this work.

## III. METHODS

### A. RBM embedding within the D-Wave lattice

The architecture used in the D-Wave hardware is the so-called Chimera graph shown in Fig. 1(a) and (b). The quantum bits (qubits) are located at vertices of the Chimera graph. Some of the qubits are connected with couplings having weights $J_{ij}$. In addition, the value of the optimization variable is influence by the bias fields $h_j$. The values of $J_{ij}$ and $h_j$ are assigned to represent a particular spin glass problem described by the following expression for N qubits:

$$E(s) = -\sum_{i=1}^{N-1}\sum_{j=i+1}^{N} J_{ij} s_i s_j - \sum_{j=1}^{N} h_j s_j \qquad (4)$$

The Chimera lattice provides at most six connections for each qubit, or less when some qubit neighbors or couplings are missing in the particular version of the hardware. A common approach to increasing the connectivity of the graph is to

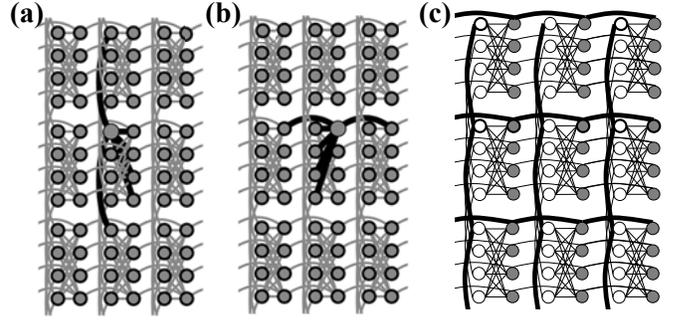

Fig. 1 (a) A magnified view of D-Wave lattice. The six available couplings for a qubit in the left column of the unit cell are highlighted with the bold lines. (b) The same as (a) highlighting the couplings for a qubit in the right column of the unit cell. (c) The RBM embedding into the Chimera lattice used in this work. The bold lines highlight "ferromagnetic" couplings that combine (clone) qubits into a single logical RBM unit in order to increase the number of connections for each RBM unit.

combine (clone) multiple qubits to represent one logical unit (Fig. 1(c)). In this case, the number of connections for the given logical unit is equal to the sum of the connections for each qubit cloned into the given unit. The cloning is conducted by assigning the maximum allowed value to the couplings $J_{ij}$ to ensure that all (or at least the majority) of the cloned qubits have the same value. Those couplings approximate the so-called ferromagnetic bonds and are shown with bold lines in Fig.1(c). The vertical bold lines in Fig. 1(c) clone qubits into visible RBM units, while hidden units are achieved by combining qubits using bold horizontal lines. For clarity of the figure, only some of the qubits are shown to be combined with bold lines.

In contrast to our previous work [15], the lattice of 16×16 unit cells of the D-Wave hardware used in this work had no missing qubits or couplings. The same embedding approach enabled 64 (16×4) visible and 64 hidden RBM units, each connected to all units of the opposite kind. Therefore, this version of the hardware allowed embedding of a complete RBM architecture.

### B. RBM training by classical CD

As described in the previous section, the price for the complete RBM connectivity is a relatively small number of RBM units that can be embedded within the D-Wave hardware. For a similar RBM embedding, a toy problem of 8×8 bars-and-stripes (BAS) [15][16] was selected in our previous work. That previously used dataset was employed also at the first stage of this work for comparison. The main focus of the current investigation, however, is the much more realistic dataset, a popular in ML OptDigits data set, which was scaled down to 8x7 pixels, binarized, and converted from the QUBO {0,1} to Ising {-1,1} format. The optimal embedding provided us with 64 visible units, which restricted us to using not all 10 but only 8 RBM units for labels. Therefore, 8 handwritten digit classes (from 0 to 7) were used for the training and reconstruction. The RBM was trained with 1024 different training patterns by a classical algorithm (without the D-Wave) and then embedded into the D-Wave lattice. Each of the 1024 training patterns was used to initiate a five-step Gibbs chain. Patterns obtained at the end of the Gibbs chain provided a sample from the model distribution during the contrastive divergence (CD) training [26]. Following an approach demonstrated in the previous work to be very effective, the values for the weight-decay parameter and the power term in the weight-decay expression were optimized by trial and error and applied to keep the absolute value of the maximum weights below 0.5, while also discouraging too small values of weights.

Bigger weights in excess of 0.5 could be too close to 1, which is the maximum valued allowed by the D-Wave hardware. The value of 1 was used in the RBM embedding into the D-Wave to establish a "ferromagnetic" bond between qubits combined to represent a single RBM unit. If some RBM weights also approach this value after training, some of the "ferromagnetic" bonds might be violated, compromising the validity of the embedding. On the other hand, too small values for $w_{nm}$ were discouraged to avoid the risk of having smaller weights dropping below the D-Wave sensitivity limit after applying an additional scaling with a D-Wave scaling factor (Fig.2).

*C. Reconstruction of incomplete images and missing labels using the RBM embedding within the D-Wave lattice*

After the RBM training using the classical CD, the D-Wave and MCMC were applied independently to reconstruct missing portions of incomplete input images and the missing labels. The procedure developed in the previous work [15] was used. For each incomplete test image drawn from the test set of 440 handwritten digits, all visible units corresponding to the available portion of the image were "clamped" to the value of the corresponding pixel. The clamping in the D-Wave hardware was achieved by providing the maximum value (-2 or 2) to the bias field $h_j$ of the corresponding qubits, depending on the value at the pixel of the test image.

The D-Wave GS was sought as a representation of the most energetically favorable combination of the remaining qubits (and therefore the corresponding visible units and labels) for the given clamped incomplete input image. A critical advantage of using the D-Wave is that the standard MCMC image reconstruction method takes multiple MCMC steps to obtain the reconstructed image and/or label. In contrast, one call to the D-Wave machine provides the immediate result for the image reconstruction and classification attempt.

From the 1000 solution repetitions provided by the D-Wave in a single call, the lowest-energy state was used. For the fixed values of the qubits corresponding to visible RBM units, this state gives the highest probability to the qubits representing the sought missing visible RBM units and labels. From this state, a majority vote was taken among the values of the qubits combined to represent the reconstructed RBM units and the labels. The obtained majority vote represented the sought values for the units and labels.

The values of $J_{ij}$ obtained from the values of $w_{nm}$ may be not sufficiently smaller in magnitude than the -1 value of the ferromagnetic couplings [11]. This lack of an overwhelming comparative strength of some ferromagnetic bonds may cause violation of the requirement that all qubits combined to represent a given RBM unit must have the same values. As a result, the lowest energy state found by the D-Wave would be the GS of a different graph, not the lowest-energy state of the RBM.

In order to further mitigate this problem, the values of $J_{ij}$ and $h_j$ were further reduced using a scale factor, which was experimentally adjusted to achieve the best results in terms of both (1) the classification error and (2) the correspondence between the GSs found by the D-Wave and the classical SA [11]. The results of optimizing the scaling factor to obtain from the D-WAve the lowest classification error for the 440 handwritten digits are shown in Fig.2. Reasonably large values of the scale factor allow keeping the values of D-Wave couplings $J_{ij}$ sufficiently below 1. As a result, most of the combined qubits are ensured to have the same value, providing a correct embedding and, as a result, an optimal for the given RBM training classification error by the D-Wave.

However, too large of a value of the scale factor may lead to the smallest values of the coupling dropping below the sensitivity limit of the D-Wave hardware, which hurts the quality of the embedding and causes the deterioration of the classification error.

*D. Determination of the local valleys in the energy function*

A conservative approach to assess if the D-Wave has a potential to offer important advantages to sampling from the RBM model distribution was introduced in Ref.[15]. The approach is further applied in this work to the RBM trained with the handwritten digit dataset, while a comprehensive statistical analysis was introduced to provide a meaningful comparison of the D-Wave and the Gibbs samples. The approach is based on comparing the D-Wave and the Gibbs samples from the point of view of what local valleys the states in the samples belong to. Further, the relative importance of the local valleys reflected in the D-Wave and the Gibbs samples was analyzed by comparing the RBM energy of the corresponding local minima, the depth of the local valleys (i.e., the height of the escape barrier), and a width-related parameter of the valley.

The classical search for the local minima of the RBM model distribution utilized Gibbs sampling. It was performed on the original RBM graph free from the representation of the RBM units by combining qubits. Each of the 1024 training patterns was used as the initial state in the Gibbs chain. A specified number of Gibbs sampling steps was conducted at an initial temperature of $T=1$. Following that, the value of $T$ was set to zero in order to perform relaxation to the bottom of the current local valley. Up to 1000 Monte Carlo sweeps were performed, until no new low-energy states were generated at the end of the relaxation at $T=0$.

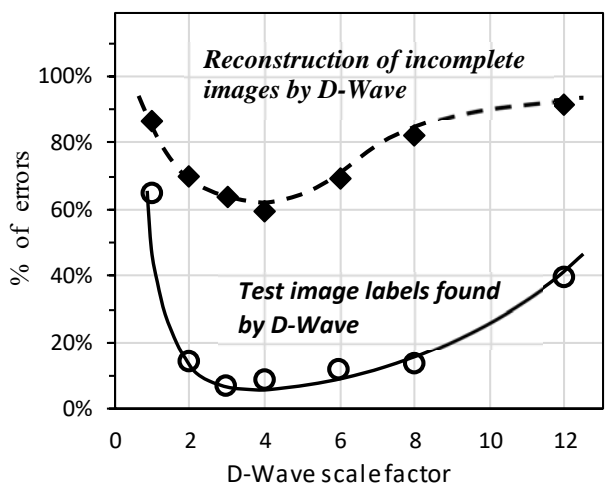

Fig. 2 The percentage of the classification errors for the handwritten digits (bottom) and the percentage of the errors of incomplete image reconstruction (top) achieved by the D-Wave, shown as a function of the scale factor applied to the couplings and bias fields of the RBM imbedding into the D-Wave.

The D-Wave was run 10,000 times in each D-Wave call to generate 10,000 attempts to find the lowest-energy solutions for the embedding of the trained RBM (described in Section III.A). The 10,000 D-Wave anneals returned 10,000 solution attempts. Some of the 10,000 solutions are found multiple times, thereby reducing the number of distinct samples from the given D-Wave call. Each distinct D-Wave solution was used as an initial state (i.e., the vector of the visible units), and a MCMC was ran for each of those states, followed by a relaxation (a SA at $T=0$) to the bottom of the local valley. The parameters of the MCMC and relaxation sequence were the same as in the case of the local minima determination using Gibbs sampling described above.

*E. Simulated Warming*

Simulated warming (SW) was implemented by conducting MCMC while gradually increasing the temperature from the initial value of $T=0$. In this version of the SW, the samples were collected and evaluated after each MCMC jump. A big enough number of the samples that remain in the same local valley could be used as a comparative estimate of the size of the local valley and, indirectly, the density of states. To include only those sampled states that remain in their initial local valley after SW, a relaxation from each of the sampled states (a SA at $T=0$) was conducted. The minimum of the local valley to which the sampled state belonged was found, which was used to establish if the local minimum indeed corresponds to the initial local valley of the D-Wave state, or if it escaped into another local valley. SW at different temperatures was also used to find the escape frequency and its activation energy $E_{act}$. The escape frequency for a given local valley was estimated by counting and taking an inverse of the number of MCMC jumps before the state finds itself outside of the local valley under consideration

## IV. RESULTS AND DISCUSSION

Classification, reconstruction and pattern generation results reported in the first part of this section primarily aim at verifying the D-Waves's ability to correctly determine the lowest energy state of a complex probability distribution under a variety of constrains (e.g., clamped visible units, clamped labels, etc). Having this reassurance about the validity of the embedding, the main results of this work are reported in Section IV, parts B and C.

*A. Using D-Wave for classification on RBM graph trained by classical CD*

Fig. 3 shows the classification error versus the training epoch, obtained (A) using the classical MCMC and (B) using the D-Wave reconstructing the qubits corresponding to the unknown labels of the test image. Here, the term epoch means a full cycle of iterations, with each training pattern participating only once. At any training iteration, the D-Wave gives at least 2 times better classification error, which was suggested by us in Ref.[15] to be due to a lower chance of the labels getting stuck in a wrong local minimum during the reconstruction of the highest-probability state. The advantage of the more than 2 times improvement of the classification error for the toy BAS dataset in Ref.[15] is reproduced here for the more difficult task of classifying 8 classes of handwritten digits.

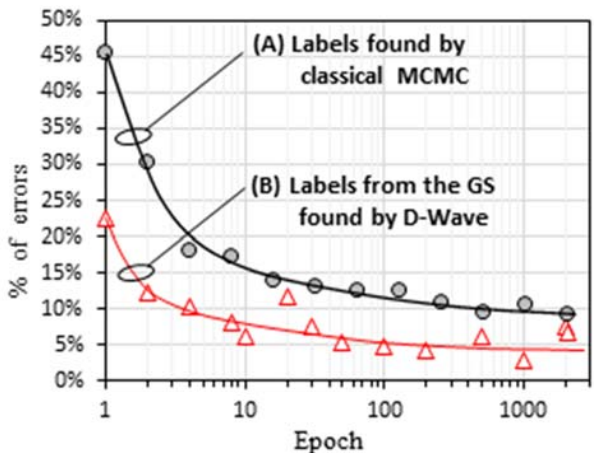

Fig. 3 The classification error versus the training epoch, obtained (A) using the classical MCMC and (B) using the label reconstruction by the D-Wave.

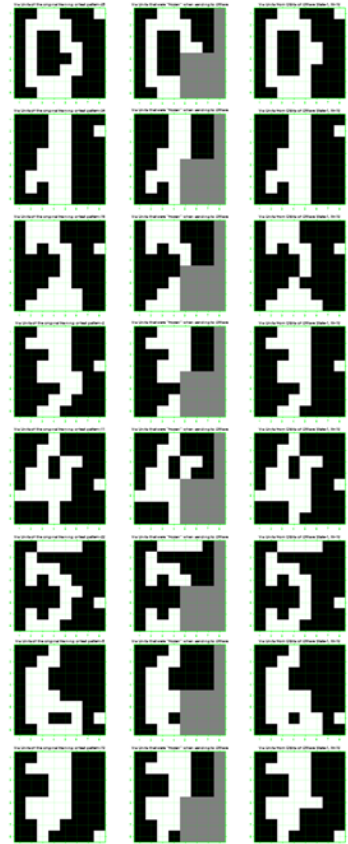

Fig. 4 Examples of correct classifications of incomplete input images of various handwritten digits using the D-Wave, after 10 training epochs. Left: the original test images. Center: input images without labels "clamped" in the D-Wave call. Right: the reconstructions (i.e., the GSs) found by the D-Wave for the missing units and labels. The rows represent examples for each of the digit classes recognized by the D-Wave.

A few examples of correct classifications of incomplete input images of various handwritten digits using the D-Wave are shown in Fig.4. The classification was conducted after only 10 training epochs.

*B. D-Wave embedding of the classically trained RBM used as a generative model.*

Further, the RBM was fully trained with 2000 epochs on the handwritten digit dataset. Only the qubits representing the labels (i.e. combined with "ferromagnetic" bonds to represent RBM units of a label) were clamped to the desirable values of the label. The 1,000 solution repetitions returned by the D-Wave provided 1,000 generated images, some of which may repeat. The five lowest-energy solutions for each of the 8 labels are shown in Fig. 5 as examples.

*C. Comparison of D-Wave and Gibbs samples from the RBM model distribution*

Similar to our results in Ref.[15], the D-Wave sample from the BAS-trained RBM, when using the embedding into the D-Wave lattice with no missing qubits and couplings, represents a smaller number of the local valleys compared to the classical sample (Fig.6 (a)). This fact alone indicates that the sample may be not sufficiently representative. Furthermore, more than 98% of the classically found local valleys are missed by D-Wave for the BAS dataset (Fig.6(b)). On the other hand, this also means that the D-Wave sample includes local valleys that were not present in the classical sample.

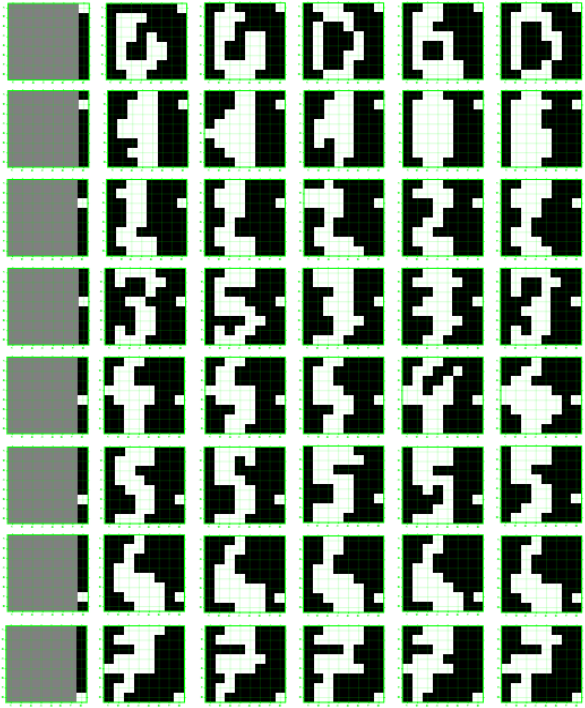

Fig. 5 Examples of using RBM embedding into the D-Wave as a generative model. The rows represent examples of images generated for eight different clamped labels. The first column shows the labels clamped during the image generation. The other five columns show five examples of the generated images (five lowest-energy states found by the D-Wave for the given value of the clamped label).

For the handwritten digit dataset, the D-Wave sample represented larger number of the local valleys than the number of the local valleys in the classical sample (Fig. 7(a)). It could have been expected from these results that the set of the local valleys present in the D-Wave sample should include at least all the local valleys from the Gibbs sample plus additional local valleys that were not present in the Gibbs sample. The reality was more disappointing. There is only a relatively small overlap between the two sets of the local valleys. For the case of the handwritten digit dataset, ~60-80% of the classically found local valleys are missed by the D-Wave (Fig.7(b)). On the other hand, it also means that the D-Wave sample included even more (compared to the BAS case) local valleys that were missed by the classical sample.

Out further analysis targeted exclusively the handwritten digits dataset as a more realistic example of a useful training case.

During the RBM training, maximizing the marginal probability (which also leads to higher joint probability and lower RBM energy) for the states having visible units corresponding to the training patterns, lower-energy states contribute the most to the partition function in Eq. (3). Obviously, a hypothetical sample from the model distribution missing local valleys with minima having low energy would be a more serious problem than missing higher-energy local valleys. The next step in the investigation aimed at establishing what kind of local valleys present in the classical MCMC sample are missed by the D-Wave sample with respect to the energy.

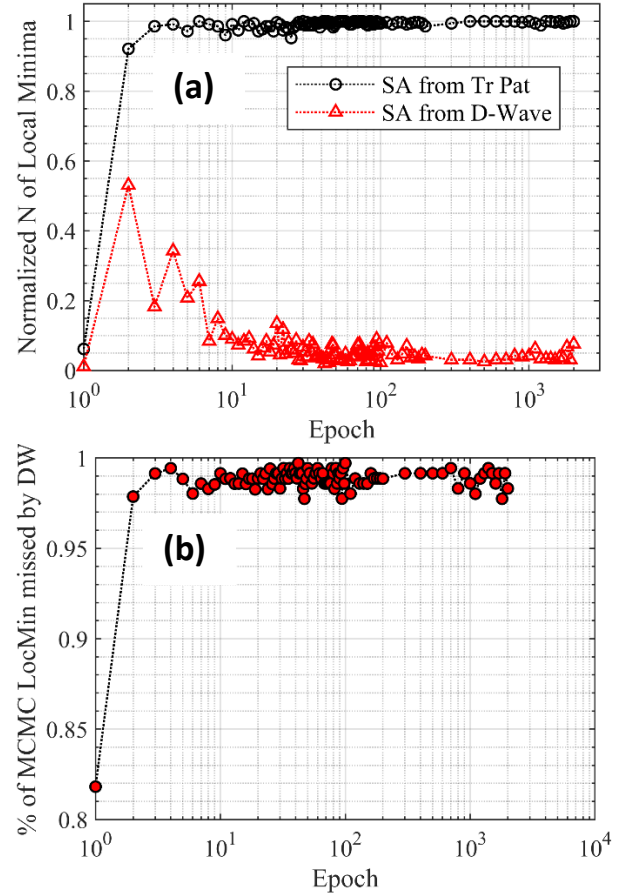

Fig. 6 RBM trained with BAS. **(a)** The total number of the found local minima as a function of the training epoch. The two dependencies in (a) correspond to (bottom) the local minima found by SA at $T=0$ from each of the 1000 repetitions of the D-Wave solutions, and (top) the local minima found by SA at $T=0$ after one classical Gibbs step from each of the training patterns. The number of the found local minima is normalized to the total number of the training patterns. **(b)** The percentage of the local valleys in (a) found by the classical MCMC that are missed by the D-Wave sample.

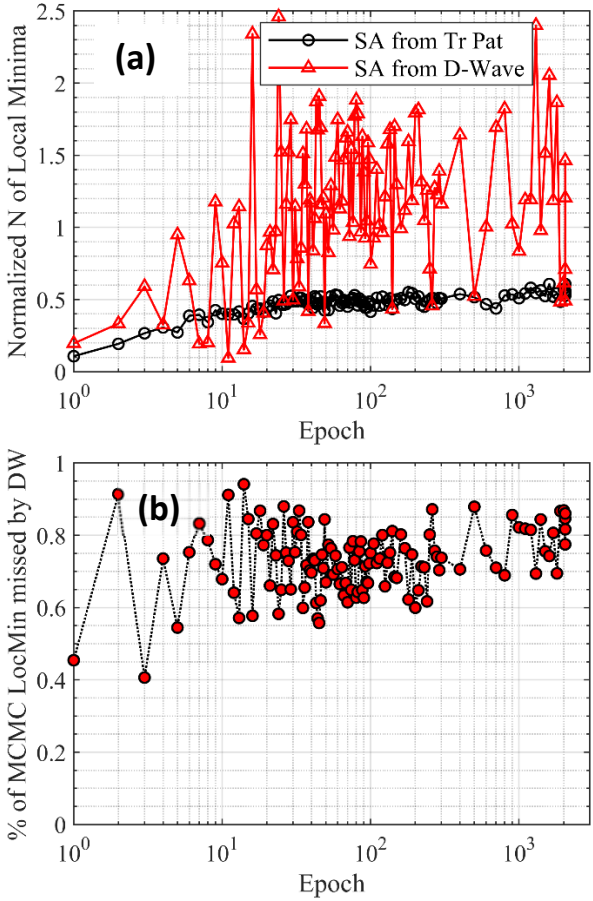

Fig. 7 RBM trained with handwritten digits. **(a)** The total number of the found local minima as a function of the training epoch. The two dependencies in (a) correspond to (top) the local minima found by SA at *T=0* from each of the 1000 repetitions of the D-Wave solutions, and (bottom) the local minima found by SA at *T=0* after one classical Gibbs step from each of the training patterns. The number of the found local minima is normalized to the total number of the training patterns. **(b)** The percentage of the local valleys in (a) found by the classical MCMC that are missed by the D-Wave sample.

Histograms of the RBM energies of the minima of the local valleys present in the classical MCMC sample are shown in Fig.8 as white bars. The dark bars in Fig.8 are for the minima of those local valleys in the MCMC sample that are also represented by the D-Wave sample. The light bars are for those MCMC-found local minima that were not found in the D-Wave sample.

All the lowest-energy local minima (the highest joint probability states in the model distribution) in the classical sample are also found by the D-Wave sample. With the increase of the RBM energy (lower probability states), an increasingly higher percentage of the classically-found local minima are not present in the D-Wave sample. The result holds for different training epochs, with three different training epochs illustrated in Fig.8.

The next question was about those "new" local valleys that were found by the D-Wave but missed by the classical MCMC sample. The corresponding histograms for the RBM energies of the local valleys present in the D-Wave sample are shown in Fig.9. The lowest-energy local valleys found by D-Wave are the same as what MCMC finds. However, the main interest now is in the "new" local valleys found by the D-Wave sample but missed by the classical technique. If follows from Fig.9 that those "new" local valleys are distributed across a wide range of the RBM energy, dominating at higher energies (lower probability states) but also present in significant amount at medium-to-low energies. This result indicates that at least some of the "new" local valleys, if present in the sample, could be an important contribution to the quality of the sample, determining how good or bad the model RBM distribution approaches the target distribution.

Further insight into what kind of the local valleys are more likely to be represented by the D-Wave sample while missed in the MCMC sample was obtained by looking at the height of the lowest portion of the barrier for the local valley, which represents the lowest energy from the bottom of the local valley that must be acquired for escaping into the neighboring valley. Fig.10 shows histograms of the activation energy $E_{act}$ of the escape frequency from the D-Wave-found local valleys during the SW.

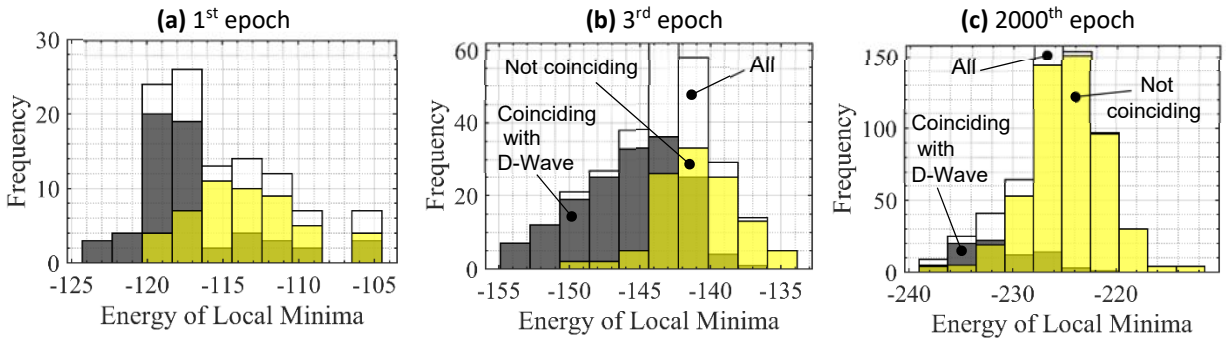

Fig. 8 Histograms of the RBM energies of the minima of the local valleys present in the classical MCMC sample (found by SA at *T=0* from each of the 1024 training patterns). Histograms are plotted for three different training epochs. The white bars are for all the local minima in the MCMC sample. The dark bars are for the minima of those local valleys in the MCMC sample that are also represented by the D-Wave sample (found by SA at *T=0* from each of the 1000 repetitions of the D-Wave solutions). The light bars are for those local minima that were not found in the D-Wave sample.

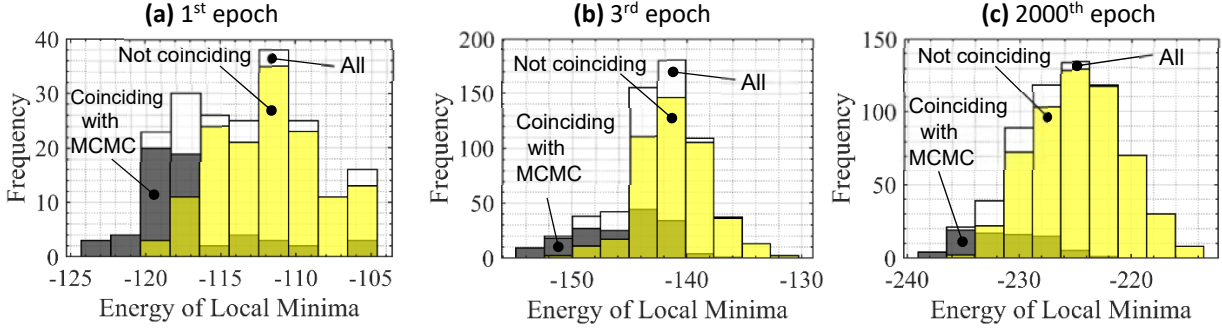

Fig. 9 Histograms of the RBM energies of the minima of the local valleys present in the D-Wave sample (found by SA at $T=0$ from each of the 1000 repetitions of the D-Wave solutions), plotted for three different training epochs. The white bars are for all the local minima in the D-Wave sample. The dark bars are for those local minima in the D-Wave sample that coincide with the local minima in the classical MCMC sample (found by SA at $T=0$ from each of the training patterns). The light bars are for those local minima that are new and not present in the classical sample.

While different heights of the escape barrier dominate at high versus low temperatures, $E_{act}$ used in (a) reflects the lowest portion of the escape barrier found from the lower-temperatures part of the Arrhenius curve. This estimate of the barrier height was independently confirmed by sampling the energy of the last state before escaping the barrier.

Four of the D-Wave-found local valleys in Fig.10 having the highest $E_{act}$ have been also found in the classical MCMC sample. Except for those four, the "new" local valleys found by the D-Wave but missed by the MCMC sample have the escape barrier height in a wide range of energies. In other words, among the "new" local valleys, there are valleys of the variety of depths from very shallow (i.e, less important for the quality of the sample) to very deep (i.e., more important).

Finally, histograms of a width-related parameter of the D-Wave-found local valleys are shown in Fig.11. The width-related parameter was estimated using an approximation of a local valley with a square well, with the width calculated by dividing the total number of the states having energy below $E_{act}$ of this valley by the depth of the well $E_{act}$. A qualitatively similar result was obtained using the intercept of the Arrhenius plot with the vertical axis as the depth estimate.

With respect to the width of the local minima, there is no clear trend for the "new" local valleys found by the D-Wave while missed by the MCMC. Among the "new" local valleys, there are valleys of the variety of width from very narrow to very wide. However, at least for this epoch, the fraction of "new" local valleys missed by MCMC seems to somewhat increase for higher local valley width. The four widest D-Wave-found local valleys happened to be all missed by MCMC.

Understanding what distinguishes the local valley found by D-Wave but missed by classical MCMC is beyond the scope of this paper. However, an important conclusion can be suggested from Figs. 8-10. It appears that the D-Wave's sample reflects some important local valleys missed by MCMC, many of which contain high-probability states, high density of states and high barrier for escaping from the local valley.

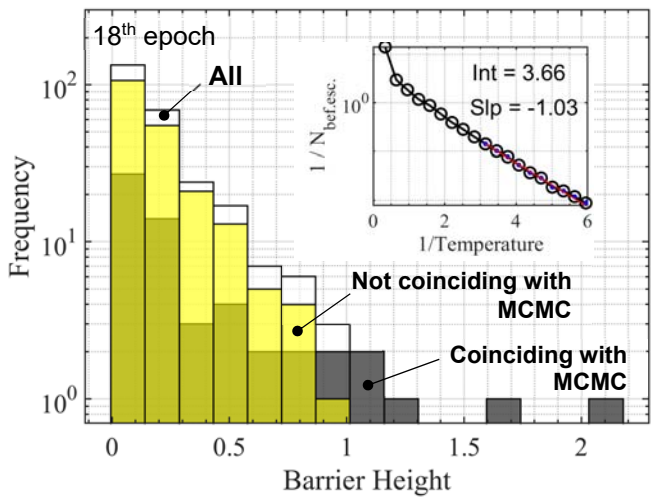

Fig. 10 Histograms of the activation energy $E_{act}$ of the escape frequency from the D-Wave-found local valleys. The dark bars are for those local valleys in the D-Wave sample that coincide with the local valleys in the classical MCMC sample. The light bars are for those local valleys that are new and not present in the classical sample. The inset in (a) illustrates the Arrhenius plot for one of the local valleys.

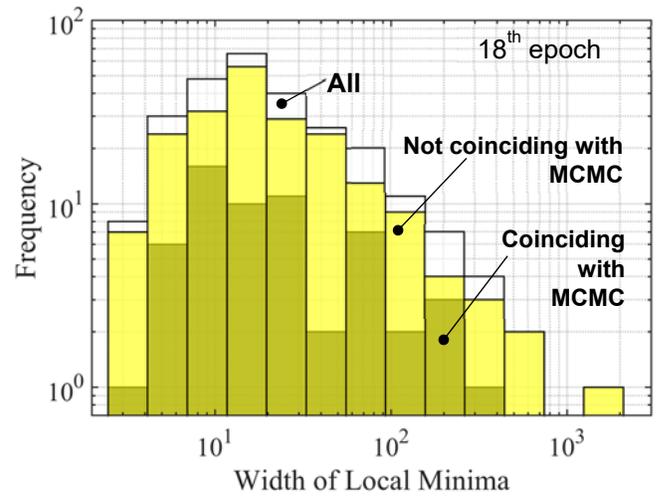

Fig. 11 Histograms of the width-related parameter of the D-Wave-found local valleys. The dark bars are for those local valleys in the D-Wave sample that coincide with the local valleys in the classical MCMC sample. The light bars are for those local valleys that are new and not present in the classical sample.

## V. CONCLUSION

Use of the D-Wave QA for classification of handwritten digits and the demonstration of the successful use of the D-Wave as a generative model is an important milestone in evaluating capabilities of QAs for Machine Learning. Both in the previous work dealing with a BAS dataset [15] and in this work that focused on a much more complex dataset, the D-Wave gives at least two times better error for incomplete image classification compared to MCMC, with much faster performance (a nearly instantaneous reconstruction of the missing visible units and labels).

Analysis of QA and MCMC samples from the RBM model distribution was conducted while using D-Wave hardware with no missing qubits and couplings. It allowed us to eliminate the concern that the (usual) imperfections of the QA hardware could be responsible for some of the trends in Ref.[15].

Statistical analysis of the distribution of the "new" local valleys represented in the D-Wave sample but missed by the classical MCMC sample indicated that the D-Wave sample contains some important local valleys often missed by classical MCMC, many of which include high-probability states, high density of states and high barrier for escaping from the local valley.

The fact of the D-Wave also missing many important local valleys present in the classical sample suggests that further improvements are needed before it is possible to fully replace MCMC with a D-Wave sampling. However, we believe that the obtained results point out at a strong potential of achieving drastic improvements in the sampling speed as well as producing a much better sample for the gradient-descent based log-likelihood maximization that would be more representative of the model distribution compared to samples obtained with MCMC, Gibbs sampling, etc. Such improvements could further empower the broad family of generative models based on nondirected probabilistic graphs, including deep architectures such as Deep Boltzmann Machines. Investigation of the training using a combined Gibbs and D-Wave sample is the subject of our future work.

ACKNOWLEDGMENT

The authors thank D-Wave Systems for access to their 2000 Q machine.

**Yaroslav Koshka** received his B.S. and M.S. in Electronics in 1993 from Kiev Polytechnic University, Kiev, Ukraine, and his PhD in electrical engineering in 1998 from the University of South Florida. From 1993 till 1995, Dr. Koshka worked as an Engineer Mathematician at the Institute for Problems of Material Science, Kiev, Ukraine. From 1998 till 2002, he was a postdoctoral fellow and an Assistant Research Professor at Mississippi State University (MSU), working predominantly in experimental micro- and nano-electronics. He joined the faculty at MSU in 2002. He is currently a Professor in the Department of Electrical and Computer Engineering at MSU and the director of the Emerging Materials Research Laboratory. His current main research area is quantum computations, their application to machine learning as well as to properties of electronic materials. Other research interests include semiconductor materials and device characterization, defect engineering, synthesis of wide-bandgap semiconductor materials and nanostructures, physics of semiconductor devices, and nanoelectronics.

**Mark A. Novotny** received his B.S. in 1973 from North Dakota State University, Fargo, North Dakota and Ph.D. in 1979 from Leland Stanford Junior University, Stanford, California, both in physics.

Since 2001 he has been Professor and Head of the Department of Physics and Astronomy at Mississippi State University, Mississippi State, Mississippi.

He has published more than 200 papers in refereed journals.

His research area is classical and quantum computational physics, with applications to classical and quantum properties of materials and statistical mechanics.

Prof. Novotny has the honor of being a Giles Distinguished Professor at Mississippi State University, as well as being a Fellow of the American Physical Society and a Fellow of AAAS.